%% file: main_arxiv.tex
\pdfoutput=1

\documentclass[11pt]{article}

\usepackage[]{acl}

\usepackage{times}
\usepackage{latexsym}

\usepackage[T1]{fontenc}

\usepackage[utf8]{inputenc}

\usepackage{microtype}

\usepackage{comment}
\usepackage{csquotes}
\usepackage{lipsum}
\usepackage{graphicx}
\usepackage{booktabs}
\usepackage{algorithmic}
\usepackage[]{algorithm2e}
\usepackage{xcolor}
\usepackage{enumitem}

\setlist[enumerate]{itemsep=0mm}

\title{Polling Latent Opinions: A Method for Computational Sociolinguistics Using Transformer Language Models}

\author{Philip Feldman \\
  ASRC Federal / Beltsville, MD, USA \\
  philip.feldman@asrcfederal.com \\\And
  Aaron Dant \\
  ASRC Federal / Beltsville, MD, USA \\
  aaron.dant@asrcfederal.com \AND
  James R. Foulds \\
  UMBC / Baltimore, MD USA \\
  jfoulds@umbc.edu \\\And
  Shemei Pan \\
  UMBC / Baltimore, MD USA \\
  shimei@umbc.edu 
  }

\begin{document}
\maketitle
\begin{abstract}
    Text analysis of social media for sentiment, topic analysis, and other analysis depends initially on the selection of keywords and phrases that will be used to create the research corpora. However, keywords that researchers choose may occur infrequently, leading to errors that arise from using small samples. In this paper, we use the capacity for memorization, interpolation, and extrapolation of Transformer Language Models such as the GPT series to learn the linguistic behaviors of a subgroup within larger corpora of Yelp reviews. We then use prompt-based queries to generate synthetic text that can be analyzed to produce insights into specific opinions held by the populations that the models were trained on. Once learned, more specific sentiment queries can be made of the model with high levels of accuracy when compared to traditional keyword searches. We show that even in cases where a specific keyphrase is limited or not present at all in the training corpora, the GPT is able to accurately generate large volumes of text that have the correct sentiment.
    

\end{abstract}

\input{text/introduction}
\input{text/related_work}

\input{text/methods}

\input{text/results}
\input{text/discussion}

\input{text/conclusions}

\input{text/future}
\input{text/ethics}



\end{document}

%% file: text/introduction.tex
\section{Introduction}
\label{sec:introduction}
Large-scale research involving humans is difficult, and often relies on labor-intensive mechanisms such as polling, where statistically representative populations will be surveyed using landline and cellphone interviews, web surveys, and mixed-mode techniques that combine modes. Often, participants in a survey may need to be recontacted to update responses as a result of changing events~\cite{fowler2013survey}. 

As social media has developed, many attempts have been made to determine public opinion by mining data that is available from online providers such as Twitter and Reddit, e.g.~\cite{colleoni2014echo, sloan2015tweets}. However, though social data can be analyzed in a variety of ways, it cannot replace the pollster asking about items that do not explicitly exist in the data.

Due to the the emergence of Transformer-based Language Models (TLMs) this may be ready to change. These models, such as the Generative Pre-trained Transformer (GPT) series developed by OpenAI, have been trained on millions of high-quality web pages. The model generates text (with words represented as high-dimensional vectors) as a function of an input sentence and its previous hidden states. These hidden states are in turn functions of the input sentence and the previous hidden states, etc. Since the model is not trained using any hand-crafted rules about language, it effectively learns its own set of rules for generating natural language. In doing so, it captures semantic and syntactic regularities in language that were previously hidden to traditional statistical methods.

As such, these models contain tremendous amounts of sociolinguistic information such as the patterns of speech, word choice, and dialects~\cite{meyerhoff2018introducing} that can be accessed \textit{interactively}. By using carefully constructed probes, the model can be \enquote{polled}, allowing researchers to access latent knowledge contained in the model in ways that are analogous to accessing the knowledge in a population by calling a random sample.  

Large transformer-based language models such as the GPT encode short-range and long-range statistical relationships between tokens. Tokens represent words, and clusters of tokens represent more nebulous structures that incorporate \textit{concepts} and \textit{beliefs}. A key point in understanding how these models behave are the concepts of \textit{interpolation} and \textit{extrapolation}~\cite{rytting2021leveraging}. Consider the following example using the GPT-3, where a prompt (in bold face) describing a fictional bird, a \enquote{Floober}, is used to create a contextually consistent description of this imaginary beast: 

\begin{displayquote}
    \textit{\textbf{The Floober is a large flightless bird that lives in semi-arid areas in Africa. Like its cousin the Ostrich}, the Floober is a member of the Struthionidae family.  The Floober was discovered in the early 19th century by explorers and naturalists like William Cornwallis Harris, who named them after the sound they made: "Floober". They were first described in 1821 by Thomas Burchell, the discoverer of Burchell's zebra.}
\end{displayquote}

In this example, the GPT-3 is able to \textit{interpolate} by arranging tokens in the model's information space related to African zoology based  on the attention relationship  relationship to the given prompt.  The GPT also \textit{extrapolates} from the prompt cue of \enquote{Floober} and \enquote{Ostrich} by accessing the concept of \textit{Struthionidae}, which include ostriches. These relationships are encoded as statistical dependencies among tokens, which means that when a token is missing from a query, the model can use its contextual knowledge to predict which other tokens should be included. This does not mean that the GPT-3 is foolproof. In this case, it makes a factual error by accessing tokens related Thomas Burchell (1799–1846)~\footnote{en.wikipedia.org/wiki/Thomas\_Burchell} rather than William John Burchel (1781 – 1863)\footnote{en.wikipedia.org/wiki/William\_John\_Burchell}, who was the first Westerner to describe the zebra for science.

Because of this ability to synthesize responses, language models such as GPTs can provide capabilities for capturing the human opinions and beliefs encoded in the training text that more resemble the traditional polling model. Rather than performing training data analysis (e.g., supervised classification), we can \textit{poll} the model's responses to probes. But to do this effectively requires that we develop methods to systematically reveal the relevant information captured in these models.
 
In this paper, we finetune~\cite{sun2019fine} a set of GPT-2 models on a Yelp corpora that reflect populations of users with distinctive views. We then use prompt-based queries to probe these models  to reveal insights into the biases and opinions of the users. We demonstrate how this approach can be used to produce  results more accurately than traditional keyword or keyphrase searches, particularly when data is sparse or missing.

In addition to the concepts of interpolation and extrapolation, we introduce the concept of language model \textit{memorization}, where models can be trained to incorporate repeating patterns. We incorporate this concept by introducing the technique of \textit{meta-wrapping}, which adds information to the training corpora that aids in the automated identifying of particular parts of the generated text. We further find a correlation of when the model is trained sufficiently to accurately reproduce these wrappings and the overall accuracy of the model in representing the explicit and latent information that it has been trained on.

Lastly, we provide methods for validating transformer language models in each of these contexts. We extensively study our methodology on Yelp data, where we have ground truth in the form of user-submitted stars, and discuss applications in other domains.

%% file: text/related_work.tex
\section{Related Work}
\label{sec:related}
Since the introduction of the transformer model in 2017, TLMs have become a field of study in themselves. The transformer 
uses self attention, where the model computes its own representation of its input and output~\cite{vaswani2017attention}. So far, significant research has been in increasing the performance of these models, particularly as these systems scale into the billions of parameters, e.g. ~\cite{radford2019language}. Among them, BERT~\cite{devlin2018bert} and GPT~\cite{radford2018improving} are two of the most well known TLMs used widely in boosting the performance of diverse NLP applications.

Understanding how and what kind of knowledge is stored in all those parameters is becoming a sub-field in the study of TLMs. 
Among them, ~\cite{petroni2019language} used probes that present a query to the mode as a cloze statement, where the model fills in a blank (e.g. \enquote{Twinkle twinkle \rule{.9cm}{0.15mm} star}).
Research is also being done on the creation of effective prompts. Published results show that mining-based and paraphrasing approaches can increase effectiveness in masked BERT prompts over manually created prompts~\cite{jiang2020can}. For example, mined prompts can be produced by mining phrases in the Wikipedia corpus that can be generalized as template questions such as \textit{x was born in y} and \textit{capital of x is y}. These can then be filled in using sets of subject-object pairs. Improvements using this technique can be substantial, with improvements of 60\% over manual prompts. Paraphrasing, or the simplification of a prompt using techniques such as back-translation can enhance these results further~\cite{jiang2020can}. 


Using TLMs to evaluate social data is still nascent. A study by \cite{palakodety2020mining} used BERT fine tuned on YouTube comments to gain insight into community perception of the 2019 Indian election. They created weekly corpora of comments and constructed a tracking poll based on the prompts \enquote{Vote for MASK} and \enquote{MASK will win} and then compared the probabilities for the tokens for the parties BJP/CONGRESS and candidates MODI/RAHUL. The results substantially tracked traditional polling.

Lastly, we cannot ignore the potential dangers of TLMs. OpenAI has shown that the GPT-3 can be \enquote{primed} using \enquote{few-shot learning}~\cite{brown2020language}. In their paper \textit{The radicalization risks of GPT-3 and advanced neural language models}~\cite{mcguffie2020radicalization}, the GPT-3 was primed using mass-shooter manifestos with chilling results. We will discuss these and other related issues in the ethics section.

%% file: text/methods.tex
\section{Methods}
For all the research involving finetuning, we used the Huggingface~\cite{wolf2019huggingface} 117M parameter GPT-2 model. This was done for two reasons:

\begin{enumerate}
    \item Increased speed: During the course of this study, we finetuned 48 models. We were able to finetune a model in 2-3 hours using one NVidia TITAN RTX. 
    \item Reduced carbon footprint: It is clearly possible to train larger models using more hardware in the same amount of time, but since this was a \textit{comparative} study, there was no need to add the cost and energy of spinning up a multi-GPU cloud instance.
\end{enumerate}


Our methods focus on understanding the \textit{memorization}, \textit{interpolation}, and \textit{extrapolation} behaviors of these language models. To do this, we made use of the Yelp Open Dataset\footnote{www.yelp.com/dataset}. The Yelp dataset contains reviews of different businesses by customers. It incorporates social-media-like text, locations, business names, and star reviews, which can serve as a form of ground truth for performing sentiment analysis on review text. More specifically, we created specific sets of corpora for these GPT behaviors:

\begin{itemize}
    \item \textit{Memorization -- Ratings and votes}: This corpora includes numeric information only, including stars and votes. This data was used to evaluate the \textit{Global} characteristics of the model.
    \item \textit{Interpolation -- Reviews with stars}: This corpora includes a review and the associated stars. We evaluate the star rating and its relationship to the review text in the ground truth and generated data. This is used to evaluate the \textit{Local} characteristics of the model.
    \item \textit{Extrapolation -- Masked reviews}: This corpora is trained using the same set of reviews as the previous item, only without any review that contains the phrase \enquote{vegetarian options}. It is used to compare the behavior of the model in zero-shot situations when compared to ground truth and the model trained using the masked data.
\end{itemize}

For the purposes of our research, we concentrate  on reviews of \textit{American} restaurants. At 1,795,036 reviews, this subset is more than three times larger than Italian, the next most common cuisine. This provided us with the widest spectrum of options with respect to sub-queries of ground truth.

\begin{figure}[t]
	\centering
	\fbox{\includegraphics[width=0.9\linewidth]{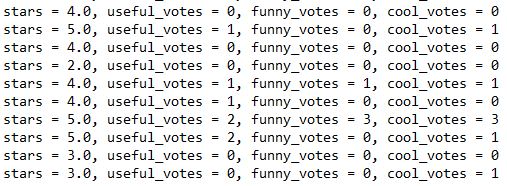}}
	\caption[]{\detokenize{Corpus section with meta-wrapping}}
	\label{fig:example_corpora}
\end{figure}

The overall technique used to create models, then generate and evaluate results is as follows:
\begin{enumerate}
    \item Download and store the Yelp dataset in a MySQL database.
    
    \item Analyze number of reviews by category. 
    
    \item Create a corpora, wrapping with meta-information (e.g. Figure~\ref{fig:example_corpora}).
    
    \item Fine-tune models, using the Huggingface API.
    
    \item Evaluate the model on a set of prompts and store the results. Each experiment contains an id, date, description, model name, list of textual probes, seed, and model hyperparameters.
    
    \item Calculate sentiment and parts-of-speech analysis on generated text\footnote{github.com/flairNLP/flair}. We also ran the same sentiment evaluation on a subset of \enquote{ground truth} reviews taken from the Yelp dataset.
    
    \item Generate charts by running queries on the database and performing analytics.
    
\end{enumerate}


We trained and evaluated three sets of models. The first sets were trained exclusively on stars and votes (See training corpora example in Figure~\ref{fig:example_corpora}). This was used to evaluate the statistical properties of the GPT against well-characterized numeric data. 

\begin{figure}[t]
	\centering
	\fbox{\includegraphics[width=0.9\linewidth]{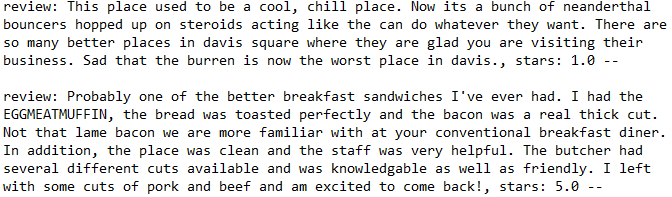}}
	\caption[]{\detokenize{yelp_review-stars_test_American_6.txt}}
	\label{fig:example_corpora2}
\end{figure}

The second sets were trained using corpora of reviews followed by stars (Figure~\ref{fig:example_corpora2}). These models were used to evaluate how effectively the models learned the relationship of the generated text to the star review. In these corpora, the training and test text were wrapped in meta information consisting of the text \enquote{review: }, \enquote{, stars: }, and terminated by a \enquote{\texttt{-{}- <CR>}}. The use of this wrapping allowed a rapid evaluation of the level of training of the model (i.e. did it learn the wrapping pattern effectively), and once learned, the meta-wrapping supported easy extraction of the synthetic data using regular expressions. 

The third set was trained using a masked corpora that did not include the phrase \enquote{vegetarian options} to compare against the other model and ground truth.

%% file: text/results.tex
\section{Results}
\label{sec:results}
In this section, we describe how the GPT is able to incorporate memorization, interpolation, and extrapolation into its behavior. We find that each one of these contexts provides useful mechanisms for determining the performance of such models. 

\subsection{Memorization}

In this section, we focus on the ability of the GPT to memorize repeating patterns while also reproducing statistically similar data with respect to ground truth. To do this, we generated \textit{meta-wrappers} from the ground truth. In this case, the number of stars, useful votes, funny votes and cool votes contained in the Yelp data. Examples of this are shown in Figure~\ref{fig:example_corpora}. When given an insufficiently large corpora, the model would fail to learn the pattern correctly resulting in generated strings like: 

\begin{displayquote}
    \small
    \texttt{\detokenize{stars_votes = 0stars_stars_stars_min = 2.0, useful_votes = 0,}}
\end{displayquote}    

However, once the corpus contained more then 50,000 lines, the model learned the pattern perfectly, and there were no more errors (Column 'error \%' in Table~\ref{tab:memorization_error_correlation}).

\begin{table}[t]
    \small
    \centering
    \begin{tabular}{lrr}
        \toprule
        model (lines) &  error \%  & correlation \%\\
        \midrule
          6k &   0.24\% & 0.36\%\\
         12k &   0.22\% & 0.62\%\\
         25k &   0.14\% & 0.86\%\\
         50k &   0.00\% & 0.96\%\\
        100k &   0.00\% & 0.99\%\\
        200k &   0.00\% & 0.98\%\\
        \bottomrule
    \end{tabular}

    \caption{Memorization Error \& Correlation}
    \label{tab:memorization_error_correlation}
\end{table}

We also tested the effects of corpus size on the ability of the model to reproduce the statistical properties of the ground truth numeric data\footnote{The vote data is mostly zeros and not as useful as the star information}. We found that increasing the number of lines improved the learning of the statistical information by the models using Pearson's correlation. However, as can be seen in the 'correlation \%' column of Table~\ref{tab:memorization_error_correlation}, it appears that the best training occurs at 50k-100k lines, with the 200k line model overfitting and  no longer generalizing~\cite{dietterich1995overfitting}.

These results indicate that the TLMs can both memorize the structure of data and  reproduce arbitrary amounts of information using that structure that are substantially similar to ground truth. These memorization properties allow us to evaluate the quality of models by injecting known ground truth into the data using meta-wrapping and evaluating the statistical properties of the results.

\subsection{Interpolation}
In this section, we explore how finetuned GPT models are able to generate data that appropriately represents the behaviors of the group that provided the corpora. In this case, we trained models on 50k and 100k review corpora using the \enquote{American} cuisine. The arrangement of the corpus used for training is shown in Figure~\ref{fig:example_corpora2}. 

\begin{figure}[t]
	\centering
	\fbox{\includegraphics[width=0.9\linewidth]{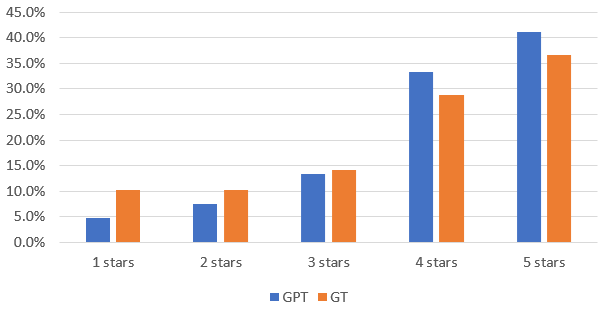}}
	\caption[]{American Ground Truth and GPT star distribution}
	\label{fig:american_gpt_gt}
\end{figure}

We then trained a model using 50k corpora and 6 epochs to compare to ground truth data. We then generated 10,000 reviews using the prompt \enquote{review:} and parsed and stored the results. Any review that ran too long to generate a star value was rejected resulting in a total of 9,228 usable review/star pairs. This model accurately reflected the distribution of stars in the ground truth with a Pearson's correlation of 99.6\% (Figure~\ref{fig:american_gpt_gt}).

An extract from a generated 4-star review is shown below:

\begin{displayquote}
\textit{\enquote{Service is good, staff is very friendly and helpful. Prices are reasonable and the restaurant is clean. The food was great. I had the veggie burger, which was great.}}
\end{displayquote}

To determine sentiment for reviews like this, we used the Flair sentiment analysis API~\cite{akbik2019flair} for each review and stored the results (6,926 positive, 2,302 negative). We also did this for 10,000 Yelp reviews selected from the \enquote{American} cuisine (6,624 positive, 3,376 negative). We then calculated the average number of stars for a POSITIVE review and a NEGATIVE review for the generated and ground truth data. The results of this comparison are shown in Table~\ref{tab:avg_star_sentiment}. 

\begin{table}[t]
    \small
    \centering
    \begin{tabular}{lrrr}
        \toprule
        Avg star rating &       GPT &        GT &  \% difference \\
        \midrule
        NEGATIVE &  2.56 &  2.29 &      5.45\% \\
        POSITIVE &  4.45 &  4.44 &      0.25\% \\
        \bottomrule
    \end{tabular}

    \caption{Star ratings for Sentiment}
    \label{tab:avg_star_sentiment}
\end{table}

The generated results are nearly identical with the ground truth, and show how well the GPT is able to generate internally consistent reviews and stars.

\begin{figure}[t]
	\centering
	\fbox{\includegraphics[width=0.9\linewidth]{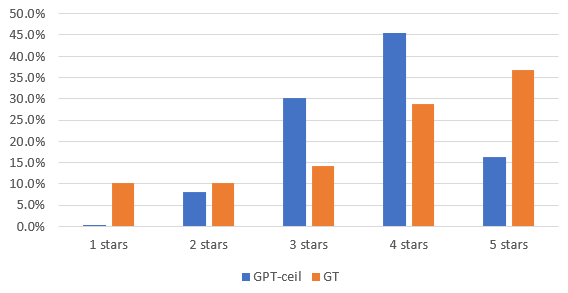}}
	\caption[]{Pre-trained GPT vs Ground Truth}
	\label{fig:pretrained_gpt_vs_gt}
\end{figure}

To see how different this was from the pretrained model, we used the prompt \enquote{What follows is a typical example of a restaurant review of an American-style taken from Yelp's database:}. This was more complex in that there was no meta wrapped output, so more complicated parsing had to be done. For instance, the GPT would sometimes rate on a 10-point rating and these scores had to be converted to the 5-point scale. Figure~\ref{fig:pretrained_gpt_vs_gt} shows a bias towards positive (4-star) reviews that is inherent in the pretrained model, while the ground truth is biased towards 5 stars. The correlation here is nowhere near the 99.6\% of the finetuned model, though it is still significant at 47.86\%.
The match of sentiment to stars is also still apparent in this data (Table~\ref{tab:pretrained_avg_star_sentiment}) even though it is less pronounced than in the finetuned GPT output. This may be partially accounted for by the ways that ratings had to be parsed and combined.

\begin{table}[t]
    \small
    \centering
    \begin{tabular}{lrrr}
        \toprule
        Avg star rating &   GPT-pre &        GT &  \% difference \\
        \midrule
        NEGATIVE &  3.25 &  2.29 &      29.46\% \\
        POSITIVE &  4.04 &  4.44 &      9.9\% \\
        \bottomrule
    \end{tabular}

    \caption{Star ratings for Sentiment (pretrained GPT-2)}
    \label{tab:pretrained_avg_star_sentiment}
\end{table}

This bias towards positive reviews in the pre-trained model may have led to  some interesting behavior on the part of the finetuned models when we tried to elicit negative (e.g. 1-star, 2-star, etc.)  reviews.  Although it was possible to produce bad reviews given a sufficiently negative prompt, the effort required to produce a one-star review was perplexing.

\begin{figure}[t]
	\centering
	\fbox{\includegraphics[width=0.9\linewidth]{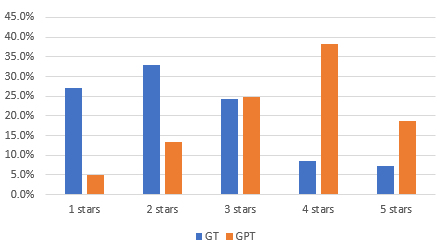}}
	\caption[]{\enquote{No vegetarian options} Unbalanced}
	\label{fig:no_veg_options_unbalanced}
\end{figure}

Figure~\ref{fig:no_veg_options_unbalanced} shows a prompt \enquote{No vegetarian options} that produced substantially negative reviews in the original reviews but produces generally positive reviews when submitted to the GPT trained on American reviews (Pearson's correlation of -63\%). 

\begin{figure}[t]
	\centering
	\fbox{\includegraphics[width=0.9\linewidth]{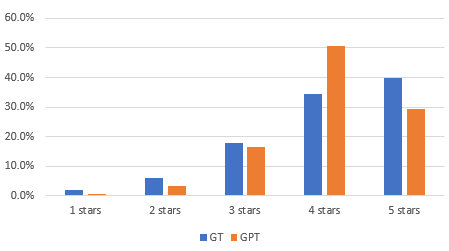}}
	\caption[]{\enquote{Many vegetarian options} Unbalanced}
	\label{fig:many_veg_options_unbalanced}
\end{figure}

Figure~\ref{fig:many_veg_options_unbalanced} shows a similar behavior for a generally positive prompt, \enquote{Many vegetarian options}. In the ground truth, there are more 5-star reviews than any other, while in synthetic reviews, the peak is again at 4 stars. This is roughly the same pattern that appears in the pretrained GPT (Figure~\ref{fig:pretrained_gpt_vs_gt}) and the negative review (Figure~\ref{fig:no_veg_options_unbalanced}).

To create an overwhelmingly one-star review with this model required the prompt \enquote{Everything about this place is terrible. The food is crap. The staff is terrible}. Clearly the model is capable of producing one-star reviews, but requires more extensive prompt tuning to do so.

\begin{figure}[t]
	\centering
	\fbox{\includegraphics[width=0.9\linewidth]{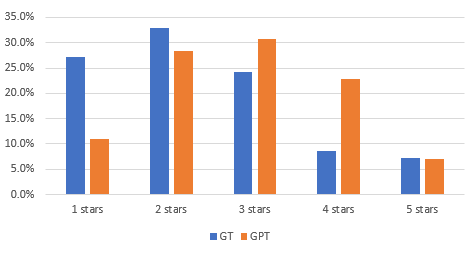}}
	\caption[]{\enquote{No vegetarian options} Balanced}
	\label{fig:no_veg_options_balanced}
\end{figure}

It appears that although there are many pathways to produce 3, 4, and 5 star reviews, there is a smaller \enquote{prompt space} that produce a sequence of tokens that produce negative reviews. Remarkably, even when the model is trained on a corpus that is \textit{balanced with respect to stars}, it still produces substantially more positive reviews for the \enquote{No vegetarian options} prompt (Figure~\ref{fig:no_veg_options_balanced}) and less 5 star reviews than the ground truth for the positive prompt \enquote{Many vegetarian options}.

To generate the appropriate sentiment/star behavior, we had to train 5 models, one for each star rating for reviews with the \enquote{American} category. Each model was trained with a 50k review corpora created from the ground truth database as shown in Figure~\ref{fig:example_corpora2}. Each model was prompted with the \detokenize{no/some/several/many vegetarian} options described above. 

\begin{figure}[t]
	\centering
	\fbox{\includegraphics[width=0.9\linewidth]{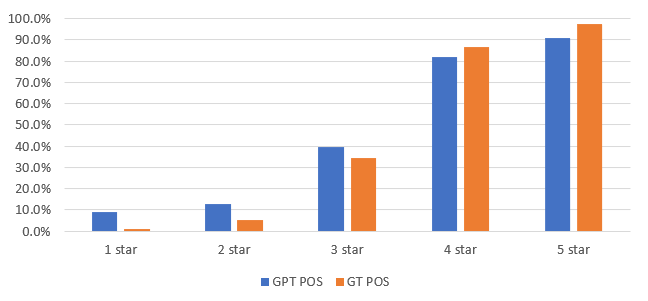}}
	\caption[]{GPT/GT Isolated Star Positive}
	\label{fig:GPT-GT_isolated_positive}
\end{figure}

\begin{figure}[t]
	\centering
	\fbox{\includegraphics[width=0.9\linewidth]{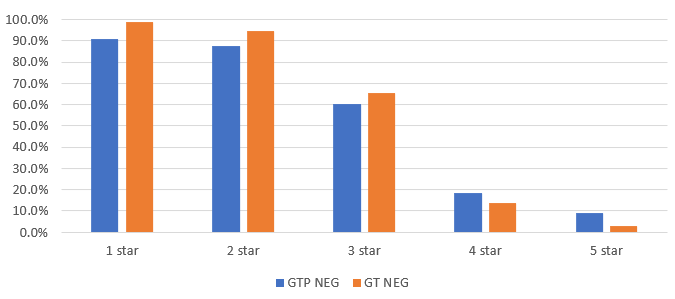}}
	\caption[]{GPT/GT Isolated Star Negative}
	\label{fig:GPT-GT_isolated_negative}
\end{figure}

The ratio of positive to negative sentiment for each model was compared to the sentiment ratio of 1, 2, 3, 4, and 5 star reviews in the ground truth data. As we can see in Figure~\ref{fig:GPT-GT_isolated_positive} and Figure~\ref{fig:GPT-GT_isolated_negative}, these correlations are much stronger (Pearson's correlation of 99.97\%) than any of the previous approaches. 

We believe that the reason that this works is because each star review represents a distinct linguistic population. On one end of the spectrum are the disgruntled, often using language that focuses on poor service such as in this extract:

\begin{displayquote}
    \textit{\enquote{We were basically seated at a table by the host, then told quite rudely by the server that we couldn't sit there.  Then we proceeded to watch as the host and server fought over whether we could sit there or not.}}
\end{displayquote}

At the other end of the spectrum is the 5-star group who have had a perfect meal with great service. These reviews are overwhelmingly classified as positive. We can show this relationship of these emotional terms to stars from a different perspective by using the Linguistic Inquiry and Word Count (LIWC) Dictionary~\cite{pennebaker2001linguistic}, which calculates the representation percentages of certain sets of words. One set of terms in the LIWC has to do with affect, ranging from positive (e.g. happy, pretty, good) to negative (e.g. hate, worthless, enemy). We can see in Table~\ref{tab:liwc_affect} how dissimilar the one and five star groups are:

    \begin{table}[t]
        \small
        \centering
        \begin{tabular}{lrr}
            \toprule
            Affect &    Pos Emo &    Neg Emo \\
            \midrule
            GPT 1 star &  2.869\% &  1.461\% \\
            GT 1 star &  2.710\% &  1.936\% \\
            GPT 5 star &  7.241\% &  0.358\% \\
            GT 5 star &  8.277\% &  0.572\% \\
            \bottomrule
        \end{tabular}

        \caption{LIWC Affect Terms for GT and GPT reviews}
        \label{tab:liwc_affect}
    \end{table}
    
These clusterings and patterns of usage allow the GPT to effectively learn the linguistic behaviors of the population so that it can accurately generate novel text that has the same sentiment patterns. And as we will see in the next section, these models are able to accurately \textit{extrapolate} text in response to prompts that do not appear in the training data, a critical element if we are to be able to use these models for polling and survey purposes.

\subsection{Extrapolation}



In our ground truth Yelp dataset, some queries result in very few reviews. When looking at only reviews with the keywords \enquote{some vegetarian options} or \enquote{no vegetarian options} there are only a handful or in the most extreme cases \textbf{no} related reviews. We can see this in the sample from the Yelp data in Tables~\ref{tab:veg_gt_pos_counts} and \ref{tab:veg_gt_neg_counts}.

    \begin{table}[t]
        \small
        \centering
        \begin{tabular}{lrrrr}
            \toprule
            POSITIVE &  no &  some &  several  &  many \\
            \midrule
            1 star &   \textbf{0} &     \textbf{0} &         \textbf{0} &     \textbf{0} \\
            2 star &   \textbf{0} &     1 &         \textbf{0} &     \textbf{0} \\
            3 star &   4 &     8 &         7 &    21 \\
            4 star &   6 &    31 &        29 &    90 \\
            5 star &   6 &    29 &        27 &   100 \\
            \bottomrule
        \end{tabular}

        \caption{Vegetarian ground truth positive review counts}
        \label{tab:veg_gt_pos_counts}
    \end{table}
    
    \begin{table}[t]
        \small
        \centering
        \begin{tabular}{lrrrr}
            \toprule
            NEGATIVE &  no &  some &  several  &  many \\
            \midrule
            1 star &  21 &     1 &         1 &     6 \\
            2 star &  24 &     6 &         8 &    18 \\
            3 star &  13 &     6 &         7 &    31 \\
            4 star &   1 &     2 &         2 &     8 \\
            5 star &   \textbf{0} &     2 &         \textbf{0} &     3 \\
            \bottomrule
        \end{tabular}

        \caption{Vegetarian ground truth negative review counts}
        \label{tab:veg_gt_neg_counts}
    \end{table}

This problem often occurs with datasets where questions may not have been asked, conditions have changed (such as the rapidly evolving information space surrounding COVID-19)  or where the structure of the data makes certain responses unlikely. This makes obtaining information about these cases difficult or impossible with traditional methods. 

Extrapolation can address this problem by letting the model extrapolate from  \enquote{adjacent} information to generate relevant, zero-shot data as we saw in the Floober example in the introduction.

To demonstrate this, we trained a new set of isolated star models on a 50k corpora that had all reviews containing the phrase \enquote{vegetarian options} \textit{removed}, or masked. These models then generated \textit{extrapolated} responses to the \enquote{no/some/several/many} prompts. 

We then compared the behavior of the \textit{interpolating} model that had been trained on corpora \enquote{vegetarian options} reviews, and a baseline of statistical samples taken from the known ground truth of 97 samples of all three-star reviews in our set of \enquote{no/some/several/many} samples. We chose baseline sample sizes of 8, 18, and 24 because those were the average size of the number of negative, positive, and combined reviews in our samples. Each sample (baseline and GPT) was randomly sampled 1,000 times and averaged for subsequent calculations. Because the GPT is able to produce unlimited reviews, we were able to use a sample size of 1,000 for these synthetic reviews.

We derived the l2 distance from POS/NEG percentage calculated from the Known Ground Truth (40.25\% / 59.74\%) for the GPT and baseline versions, which is shown in Table~\ref{tab:gt_vs_extrap_vs_baseline}. We can clearly see that the baseline(8) has the highest l2 error (20.01\%), while the GPT trained on the unmasked corpus has the lowest. Remarkably, the masked, \textit{extrapolating} GPT model has the second-lowest error, and has less than half the error of the baseline(26) evaluation.

    \begin{table}[t]
    \small
        \centering
        \begin{tabular}{llrrr}
            \toprule
             &  Pos \% &  Neg \% &  Error l2 \\
            \midrule
              Ground Truth &    40.25\% &    59.74\% &       0.00\% \\
                       GPT &    40.71\% &    59.28\% &       1.89\% \\
              GPT (no veg) &    37.58\% &    62.41\% &       3.88\% \\
              baseline(26) &    39.38\% &    60.12\% &       9.16\% \\
              baseline(18) &    40.55\% &    59.44\% &       11.78\% \\
              baseline(8) &    39.87\% &    60.12\% &       20.01\% \\
            \bottomrule
        \end{tabular}

        \caption{Ground Truth vs. Extrapolation vs. Baseline}
        \label{tab:gt_vs_extrap_vs_baseline}
    \end{table}

This is important because it demonstrates that the GPT (no veg) model is able to generate text related to vegetarian options \textit{despite being trained on data with no reviews related to vegetarian options}. These results are substantially better than the baseline even when the baseline includes over 25\% of the existing vegetarian samples. The model's ability to generate matching sentiment reviews is based purely on extrapolating between the rest of the reviews it was trained on.

These results mean that we can use language models such as the GPT to effectively learn the linguistic behaviors of the population and generate accurate responses to questions that have never been asked of the original group  but are \textit{latent} in the weights of the model. This technique creates a powerful new capability for polling and survey purposes.

%% file: text/discussion.tex
\section{Discussion}
Polling transformer language models has provided us with a new lens to assess public attitude/opinions well beyond dining options. The same technique can be used on to determine social, political and public health issues using corpora from a variety of sources. It is dynamic and can be used to answer questions using latent information. Further, is computationally inexpensive and does not require any costly human annotated  ground truth to train. 

The strength of the GPT is also a weakness. Because it stochastically generates each new token based on the ones that preceded it, but also on randomness-introducing parameters such as temperature, it can be difficult to make it behave in ways that are both predictable and dynamic. A temperature of zero will produce the same result repeatedly, but then the distribution of responses to the prompt will be lost. The best way to use these models may be to focus on the statistics of large-scale patterns rather than looking at individual responses. Stochasticity ensures that some percentage of texts will untrustworthy, but at scale such outliers can be identified and handled appropriately.

In addition, prompt design is tricky. Small changes in prompts may result significant changes in results (e.g., \enquote{some vegetarian options} versus \enquote{many vegetarian options}). Limitations of the TLMs themselves may also prevent them from providing accurate information.  For example, although humans can link affordances (\textit{I can walk inside my house}) and properties  to recover information that is often left unsaid (\textit{the house is larger than me}), TLMs struggle on such tasks~\cite{forbes2019neural}. 

TLMs are also vulnerable to \textit{negated} and \textit{misprimed} probes. Simply adding \enquote{not} to a probe (e.g. \enquote{The theory of relativity was \textit{not} developed by}, often generates \enquote{Albert Einstein}. Mispriming, or the addition of unrelated content to the prompt (e.g. \enquote{Dinosaurs? Munich is located in} the probe can produce highly distorted results.~\cite{kassner2019negated}

In this paper, we have shown that TLMs such as the GPT can be used as an effective data collection technique to gain a deeper understanding of sample populations. We believe these techniques can also be used to explore social, political and health issues, but it is important to understand their limitations.

%% file: text/conclusions.tex
\section{Conclusions}
In this paper, we described a new method for polling online data sources that uses broad keywords (e.g. cuisine = "American", stars = "3") to extract a corpora that is used to train a TLM such as the GPT. The finetuned model captures sociolinguistic patterns of the group polled that can then be accurately queried using highly targeted prompts such as "no vegetarian options".  


This unique method of querying a population on content that may not exist explicitly in the ground truth can be achieved due to TLMs capacity for memorization (learning repeating patterns), interpolation (creating variations on existing values), and extrapolation (inferring new content from existing).

We demonstrated that using TLMs in this way is actually more reliable/accurate than using ground truth queries that produce sparse results, even if the TLM model is not trained on the specific topics of interest. This opens up a tremendous opportunity for textual research where relevant data is missing, in small quantity, or volatile.

%% file: text/future.tex
\section{Future Work}

So far, we have only scratched the surface trying to  probe and understand the latent knowledge captured in a transformer language model. Our next work will involve using this technique to poll latent information on Twitter regarding public health issues. This will involve training our models on left-wing, right-wing and other groups participating in the ongoing COVID-19 online discussions. We will also be exploring the effects of negation, mispriming, and other techniques that may distort the latent knowledge captured by these models. 

%% file: text/ethics.tex
\section{Ethical Considerations}



Large Transformer Language Models' capacity to rapidly generate unethical or dangerous content (e.g. realistic mass-shooter manifestos) is well understood. Beyond the risk of the generation of credible fake content, there are additional risks for social research using TLMs.

The methods by which the latent information is stored in the model weights is a form of dimension reduction that cannot incorporate all of the nuance in the data it has trained on as it learns linguistic patterns in the data. As a result, it will inevitably fail to capture outlier behaviors in the model weights.

Even for patterns which are largely correct, the models are capable of making informational errors, such as the improper attribution demonstrated in our Floober example in Section~\ref{sec:introduction}. The model followed the highly credible linguistic pattern of an academic or Wikipedia description of an animal, complete with a likely animal family, and attributed the discovery to a person, but it was the \textit{wrong} person.

This class of error makes the latent information in TLMs valuable for population scale questions, but potentially dangerous for attributable content. The results of the model are that of generalized linguistic behaviors, and not attributable to a specific individual. Prompt tuning the model with quotes from a particular individual might provide salacious or unethical content which not only has never been produced by the individual, but includes ideas they may abhor. In fact, persistent biases or stereotypical behaviors often exist within the model's weights~\cite{abid2021persistent} \cite{nadeem2020stereoset}.

As a result, it would be extremely dangerous to utilize this sort of latent information to perform predictive actions on individuals as a result of the output of these models. AI is increasingly being applied to predictive tools for law enforcement, employment screening, and other systems that judge individuals based on an algorithmic assessment~\cite{broadhurst2019artificial} \cite{ponce2021ai}. Attempting to leverage the techniques we've demonstrated for a system of that nature would be potentially misleading, possibly dangerous, and certainly unethical.